\documentclass{bmvc2k}

\DeclareMathOperator*{\argmax}{arg\,max}
\usepackage{amsmath}
\usepackage{amssymb}
\usepackage{multirow}
\usepackage{bm}
\usepackage{floatrow}
\newfloatcommand{capbtabbox}{table}[][\FBwidth]
\usepackage{makecell}
\usepackage[ruled,linesnumbered]{algorithm2e}

\title{Noisy Annotation Refinement for\\ Object Detection}
\addauthor{Jiafeng Mao}{mao@hal.t.u-tokyo.ac.jp}{1}
\addauthor{Qing Yu}{yu@hal.t.u-tokyo.ac.jp}{1}
\addauthor{Yoko Yamakata}{yamakata@hal.t.u-tokyo.ac.jp}{1}
\addauthor{Kiyoharu Aizawa}{aizawa@hal.t.u-tokyo.ac.jp}{1}

\addinstitution{
 The University of Tokyo\\
 Tokyo, Japan
}

\runninghead{Mao, Yu, Yamakata, Aizawa}{Noisy Annotation Refinement}

\def\etal{\emph{et al}\bmvaOneDot}

\begin{document}

\maketitle
\vspace{-10pt}
\begin{abstract}
Supervised training of object detectors requires well-annotated large-scale datasets, whose production is costly. Therefore, some efforts have been made to obtain annotations in economical ways, such as cloud sourcing. However, datasets obtained by these methods tend to contain noisy annotations such as inaccurate bounding boxes and incorrect class labels. In this paper, we propose a new problem setting of training object detectors on datasets with entangled noises of annotations of class labels and bounding boxes. Our proposed method efficiently decouples the entangled noises, corrects the noisy annotations, and subsequently trains the detector using the corrected annotations. We verified the effectiveness of our proposed method and compared it with state-of-the-art methods on noisy datasets with different noise levels. The experimental results show that our proposed method significantly outperforms state-of-the-art methods.
\end{abstract}

\vspace{-15pt}
\section{Introduction}
\label{sec:intro}
Object-detection technologies using supervised learning have achieved rapid development in recent years. Most research results on object detection are based on reliable well-annotated datasets, such as ImageNet~\cite{deng2009imagenet} and MS-COCO~\cite{lin2014microsoft}. The production of a well-annotated dataset is labor intensive and extremely expensive in real-world applications, and some efforts~\cite{david2020global,papadopoulos2017extreme,xiao2015learning, kuznetsova2020open} have been made to reduce the cost of dataset production. However, the datasets obtained by these methods usually contain noisy annotations such as inaccurate bounding boxes and incorrect class labels. Owing to the interference of noise, the detectors trained on these datasets tend to perform worse than detectors trained on well-annotated datasets. 

\begin{figure*}[t]
\begin{center}
   \includegraphics[width=1\linewidth]{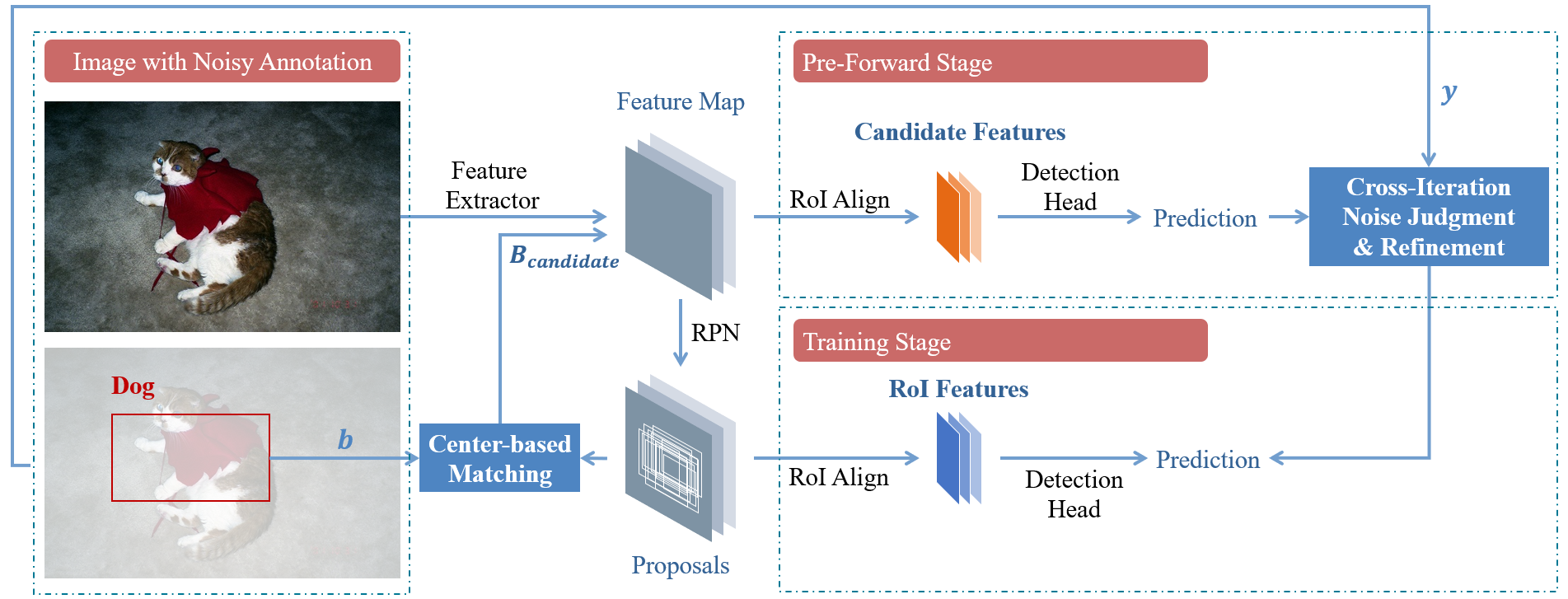}
\end{center}
   \caption{Architecture of the proposed framework. The center matching initially corrects the annotated bounding boxes and outputs the results grouped with the candidate regions, as detailed in Section~\ref{sec:CM}. The pre-forward stage classifies and regresses the candidate regions, and the cross-iteration noise judgment identifies the noisy labels on the basis of the results; these are introduced in Sections~\ref{sec:pre-forward} and \ref{sec:CINJ}, respectively.}
\label{fig:architecture}
\end{figure*}

 In this paper, we refer to datasets that contain incorrect annotations as noisy datasets. We propose a novel problem setting of training object detectors on noisy datasets with entangled noise. As multi-task learning models, object detectors are affected by incorrect annotations in a far more complex manner than CNN-based image classifiers. The entanglement of incorrect classification annotations and inaccurate localization annotations makes the analysis of complex effects challenging. For example, an incorrect localization annotation not only can directly mislead the training of the regression head of the detector but also can affect the training of the classification head by providing misplaced region features. From the results of the experiments on artificial noisy datasets, we observe a huge gap in performance between the detectors trained on noisy datasets and those trained on clean datasets. To this end, we propose a framework based on Faster R-CNN to mitigate the negative impact of both types of noise by correcting incorrect training annotations online. The main idea of our proposal is to correct the two types of noise in the dataset by the prediction of the model as the training proceeds and subsequently using the corrected annotations to supervise the training of the model. 

We summarize the contributions of this paper as follows:
\begin{itemize}
    \setlength\itemsep{0pt}
    \item We propose a novel problem setting, which is training object detector on dataset with entangled noisy annotation. 
    \item We comprehensively evaluate the impact of location noise and classification noise on object detector. 
    \item We propose a framework and several novel methods to decouple the entangled noise. By reducing the noise level of the noisy training dataset, our framework can efficiently alleviate the negative impact of noisy annotations on object detector.
    \item We conduct experiments on artificial noisy datasets with different types and levels of noise; our proposal significantly outperforms state-of-the-art methods.
\end{itemize}

\vspace{-15pt}
\section{Related Work}

For training DNNs with noisy data in classification task, Arpit \etal~\cite{arpit2017closer} suggested that DNNs first learn simple patterns and subsequently memorize noisy data. Furthermore, it was verified that noisy labels lead to high training losses~\cite{tanaka2018joint}. Based on these results, memorization-based methods for image classification have been proposed. Co-teaching \cite{sugiyama2018co,yu2019does} simultaneously trains two classifiers, and each classifier selects samples with relatively low losses in each mini-batch to teach the peer classifier. The joint optimization framework~\cite{tanaka2018joint} uses model prediction to update the labels in the dataset and uses a label distribution-based regularization term to prevent the model from irreversibly over-modifying the dataset. In addition, methods such as JoCoR~\cite{wei2020combating} and symmetric cross-entropy loss~\cite{wang2019symmetric} that use the predictions of the model to guide the training are proposed.

Unlike with the noisy label-supervised image classification, which has achieved many results~\cite{goodfellow2014explaining,jindal2016learning,zhang2017mixup,patrini2017making,sukhbaatar2014training,vahdat2017toward,zhang2016understanding}, only a few studies have focused on the noisy annotation-supervised object detection. Mao \etal~\cite{mao2020noisy} observed a positive correlation between the noise level of the dataset and the training loss of the object detector, and consequently applied the joint optimization framework to the noisy annotation-supervised object-detection task, thereby significantly reducing the impact of noise on the model accuracy. However, their method only focuses on localization annotations and assumes that all classification labels are correct. NOTE-RCNN~\cite{gao2019note} addresses the noisy annotations using a pre-trained detector, whose training requires a small number of clean annotations. Li \etal~\cite{li2020towards} used a gradient-based approach, named CA-BBC, to correct noisy localization annotations and a pseudo-label-based approach to correct noisy classification labels. They updated all annotations in the training dataset indiscriminately. In contrast to their method, our method first discriminates noisy labels using a reliable method and subsequently applies pseudo-label correction to those labels judged to be incorrect on the basis of the accurate noise judgment.

\vspace{-10pt}
\section{Proposal}
 For simplicity, in the following explanation, we assume that there is only one object and its corresponding annotation is $\{b,y\}$ on each image $x$, where $b$ denotes the annotated bounding box of the object and $y$ denotes the class label of the object, and there is only one image in each iteration. In practice, the steps described below are applied to multiple objects in each input image.
 
\vspace{-10pt}
\subsection{Pipeline}
\label{sec:pre-forward}

\textbf{Faster R-CNN} In our proposed framework, we train the Faster R-CNN with an additional pre-forward stage to decouple the entangled localization annotation noise and classification noise. Faster R-CNN consists of the feature extractor $\mathcal{E}$, the region proposal network (RPN) $\mathcal{R}$, the classification head $\mathcal{H}_c$ and the regression head $\mathcal{H}_r$. For each iteration during the training, the feature map $f$ of the training image $x$ is extracted by the feature extractor $\mathcal{E}$ as follows:
\vspace{-3pt}
\begin{equation}
f = \mathcal{E}(x).
\end{equation}

The RPN $\mathcal{R}$ takes $f$ and generates region proposals $P$, which is denoted by
\vspace{-3pt}
\begin{equation}
\label{eq:p}
P = \mathcal{R}(f).
\end{equation}

Then, $P$ is subsequently sampled and annotated on the basis of the annotation $\{b,y\}$, and each region $r$ in sampled $P$ is used to extract the region features $f_r$ as follows:
\vspace{-3pt}
\begin{equation}
f_r = \text{RoIAlign}(f, r).
\end{equation}

Finally, the classification head $\mathcal{H}_c$ takes the region features and outputs the class-wise probability $s(r)$ of each region $r$, which is denoted by:
\vspace{-3pt}
\begin{equation}
s(r) = \mathcal{H}_c(f_r).
\end{equation}

At the same time, the regression head $\mathcal{H}_r$ takes the region features as follows:
\vspace{-3pt}
\begin{equation}
r_{regressed} = \mathcal{H}_r(r, f_r),
\end{equation}
where $r_{regressed}$ is the regressed region of $r$, which is the refined region according to the offset prediction of the regression head. 

In the general training of the Faster R-CNN, the feature extractor $\mathcal{E}$, the RPN $\mathcal{R}$, the classification head $\mathcal{H}_c$, and the regression head $\mathcal{H}_r$ are optimized by SGD under the supervision of the annotated region proposals. To deal with the noisy annotation, we insert an additional pre-forward stage after the generation of the proposals $P$ (Eq.~\ref{eq:p}) and before the training of the network, as shown in Figure \ref{fig:architecture}.

\noindent\textbf{Pre-forward Stage for Annotation Refinement} First, to deal with the localization noise, we propose a center-matching correction to generate an initially corrected bounding box $b^*$ and a set of candidate regions $B_{candidate}$ based on the region proposals $P$, which will be detailed in Section~\ref{sec:CM}. Then, the region features of $b^*$ and $B_{candidate}$ are calculated using the RoI Align~\cite{he2017mask} as follows:

\begin{equation}
f_{b^*} = \text{RoIAlign}(f,b^*),
\end{equation}
\vspace{-3pt}
\begin{equation}
F_{candidate} = \{\text{RoIAlign}(f,b)~|~b \in B_{candidate}\}.
\end{equation}

Next, we feed $f_{b^*}$ and $F_{candidate}$ to the classification head and the regression head as follows:

\begin{equation}
s(b^*) = \mathcal{H}_c(f_{b^*}),
\end{equation}
\vspace{-3pt}
\begin{equation}
S_{candidate} = \{\mathcal{H}_c(f) ~|~ f \in F_{candidate} \},
\end{equation}
\vspace{-3pt}
\begin{equation}
B_{regressed} = \{\mathcal{H}_r(b,f)~|~(b,f) \in (B_{candidate}, F_{candidate})\},
\end{equation}
where $s(b^*)$ and $S_{candidate}$ denote the classification probability of the initially corrected bounding box $s(b^*)$ and candidate regions $B_{candidate}$, respectively. $B_{regressed}$ denotes the regressed candidate regions of $B_{candidate}$.

To identify the label noise, we apply cross-iteration noise judgment (as described in Section \ref{sec:CINJ}) on the basis of the classification loss $\mathcal{L}(b^*,y)$ for $b^*$, which is denoted by
\vspace{-3pt}
\begin{equation}
\mathcal{L}(b^*,y) = \text{CrossEntropy}(s(b^*), y).
\end{equation}

For each label $y$ that is judged to be noisy, we use the class having the largest predicted probability as the pseudo-label $y^*$, which is denoted by
\vspace{-2pt}
\begin{equation}
\label{eq:pseudo_1}
    y^* = \argmax_{c}{s(c|b^*)},
\end{equation}
where $c$ denotes the classes of the dataset. When 
\vspace{-3pt}
\begin{equation}
\label{eq:pseudo_2}
    s(y^*|b^*) > 0.5,
\end{equation}
we consider the pseudo-label $y^*$ to be reliable and we update $y$ by $y^*$. 
Otherwise, $\{y, b\}$ is discarded and not used for training in the current iteration. Once an annotation $\{y^*, b\}$ is used for training, the annotated bounding box $b$ is further modified to $b_m^*$ as follows:
\vspace{-3pt}
\begin{equation}
\label{eq:step_2}
    b_m^* = (b^* + b_1^* + b_2^*)/3,
\end{equation}
where $\{b_1^*,b_2^*\} \subset B_{regressed}$ denotes the regression results of $\{b_1,b_2\} \subset B_{candidate}$, where $\{b_1, b_2\}$ are the candidate regions with the highest and second-highest confidence in category $y^*$. We use the refined annotation $\{y^*,b^*_m\}$ to train the Faster R-CNN.

\begin{figure}[t]
\begin{minipage}[b]{0.24\linewidth}
  \centering
  \centerline{\includegraphics[width=\linewidth]{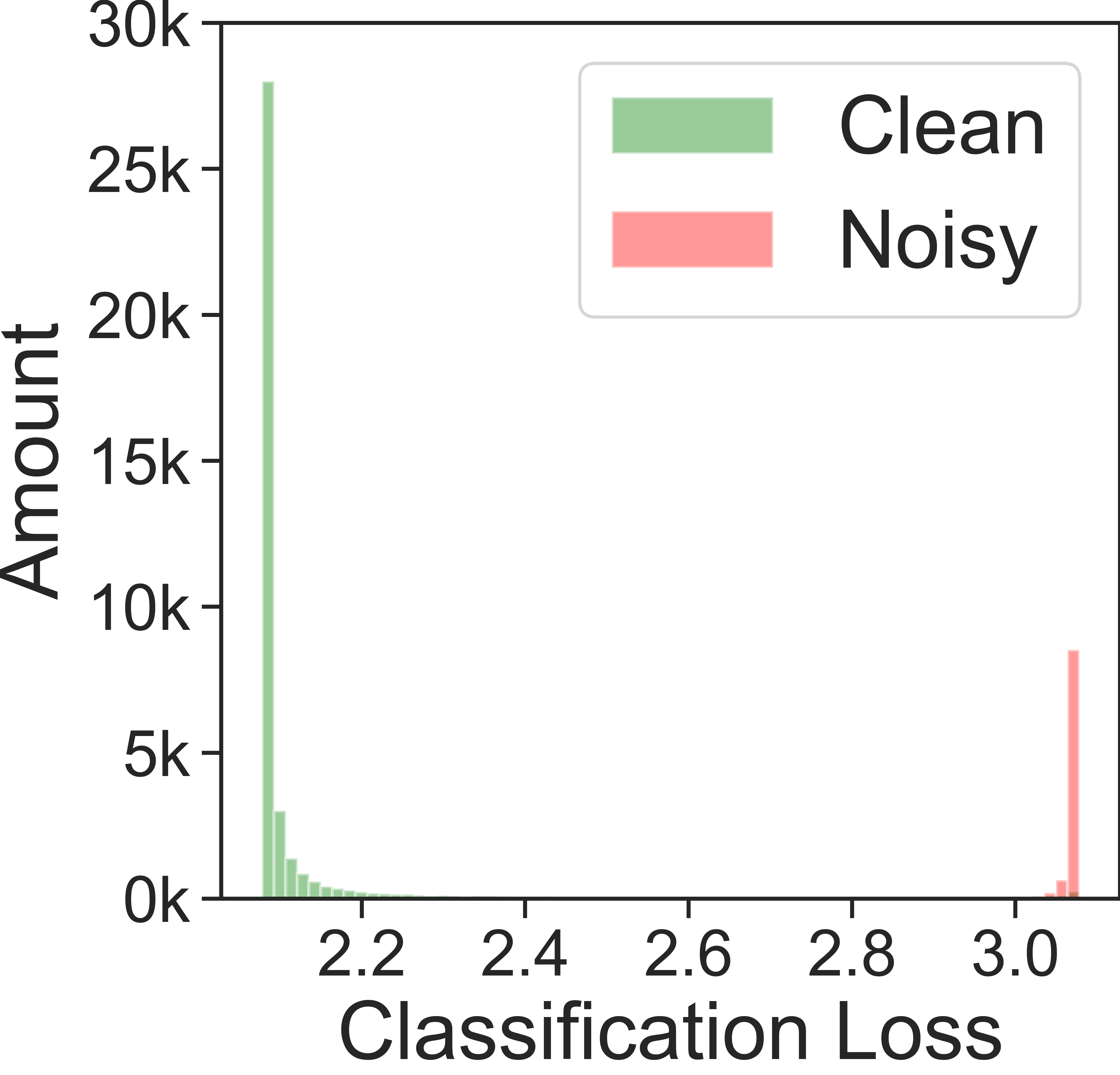}}
  \centerline{$N_{Label} = 20\%$}\medskip
  \centerline{$N_{BBox} = 0\%$}
\end{minipage}
\hfill
\begin{minipage}[b]{0.24\linewidth}
  \centering
  \centerline{\includegraphics[width=\linewidth]{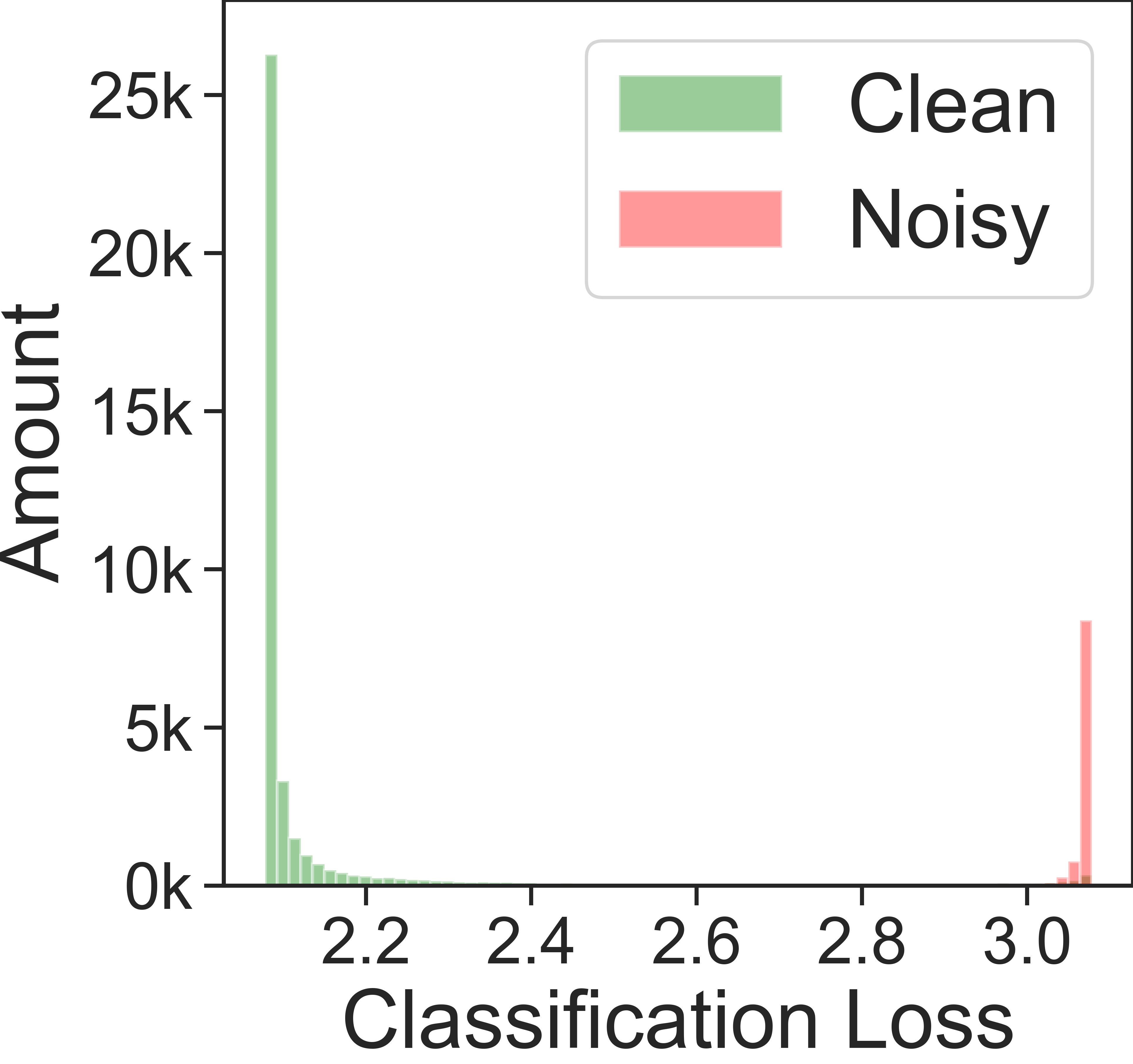}}
  \centerline{$N_{Label} = 20\%$}\medskip
  \centerline{$N_{BBox} = 40\%$}
\end{minipage}
\hfill
\begin{minipage}[b]{0.24\linewidth}
  \centering
  \centerline{\includegraphics[width=\linewidth]{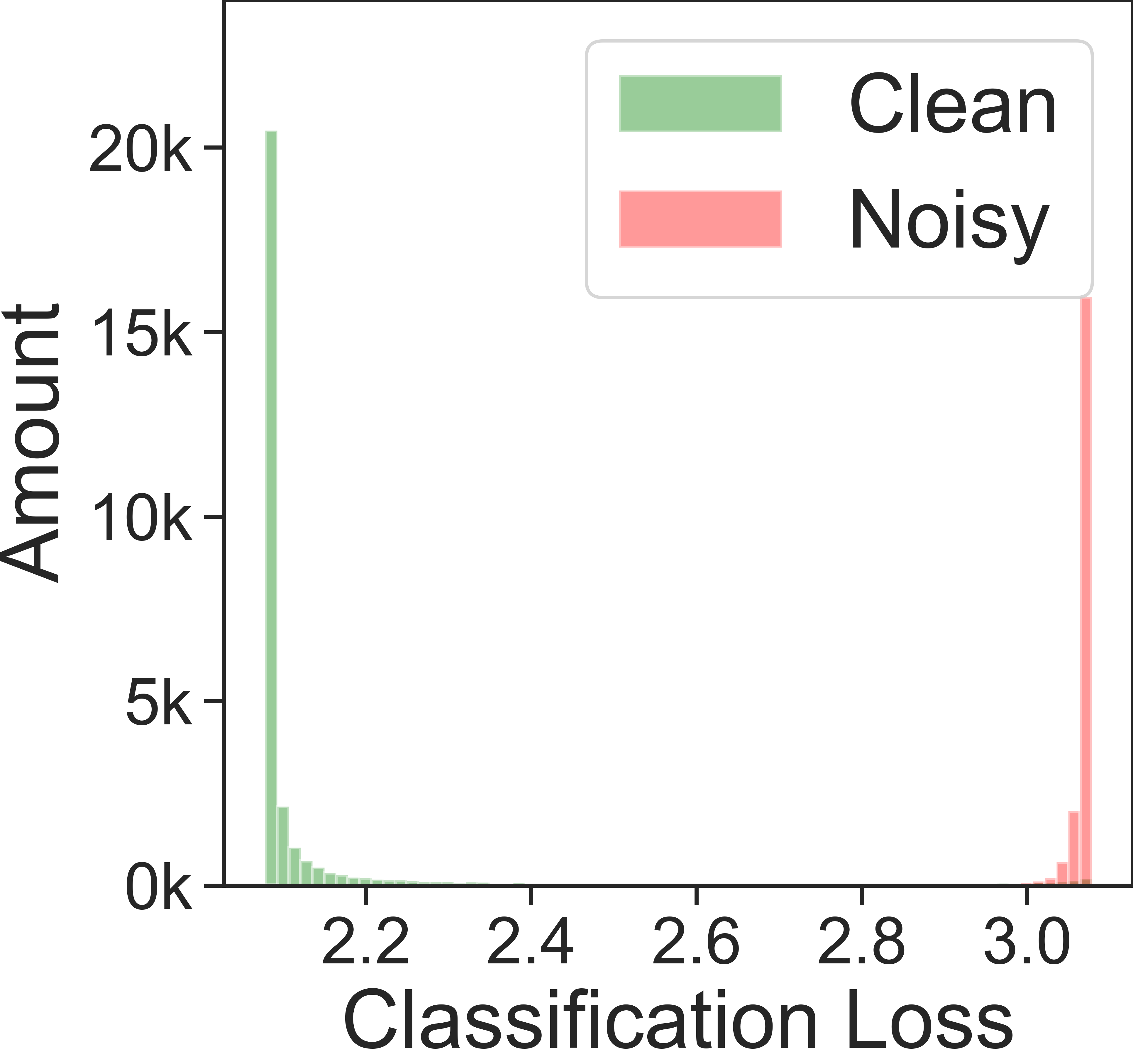}}
  \centerline{$N_{Label} = 40\%$}\medskip
  \centerline{$N_{BBox} = 0\%$}
\end{minipage}
\hfill
\begin{minipage}[b]{0.24\linewidth}
  \centering
  \centerline{\includegraphics[width=\linewidth]{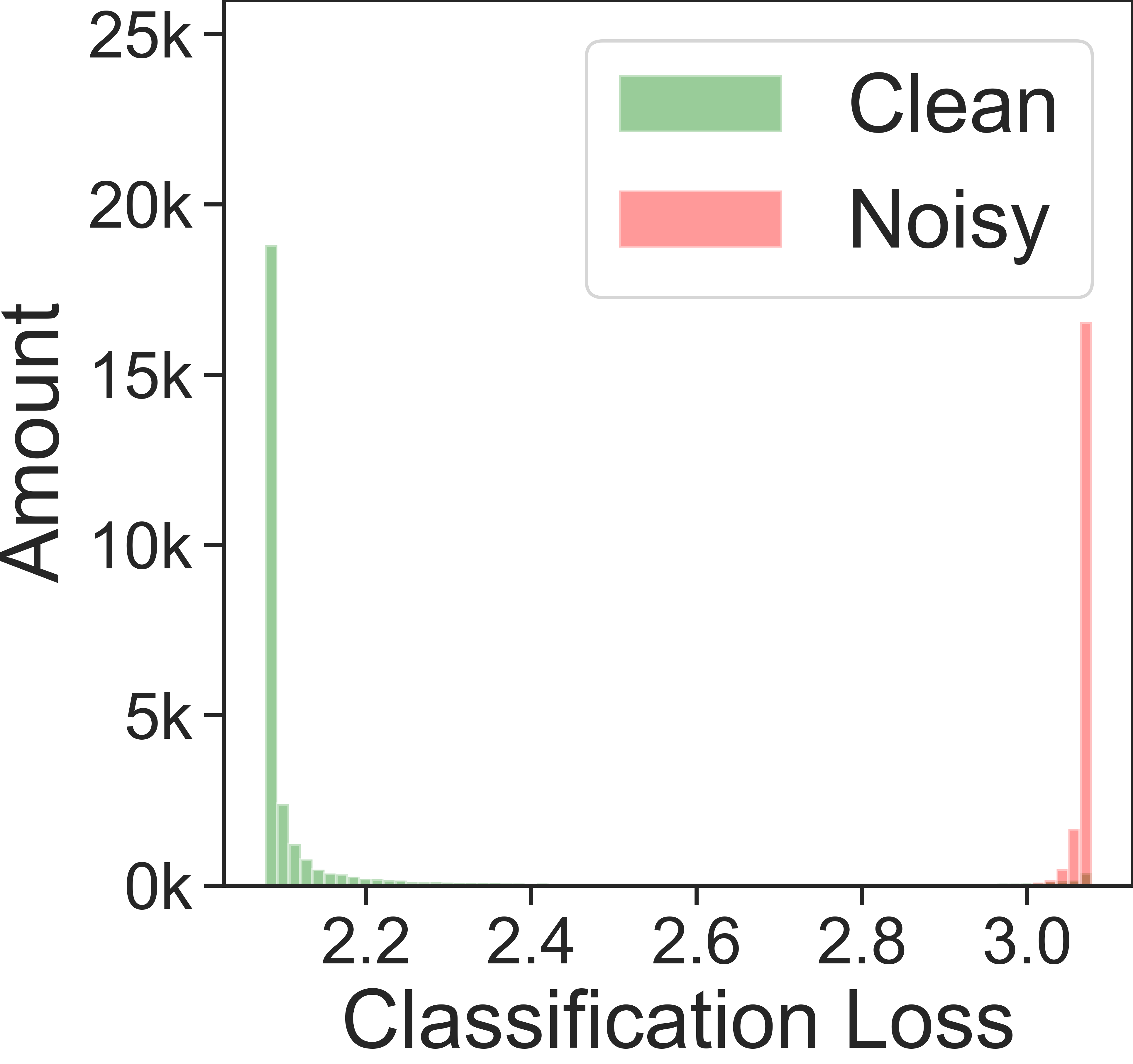}}
  \centerline{$N_{Label} = 40\% $}\medskip
  \centerline{$N_{BBox} = 40\%$}
\end{minipage}

\caption{Distribution of the classification loss caused by noisy labels and clean labels. The loss values of the noisy labels are clearly separated from those of the clean labels.}
\label{fig:distribution}
\vspace{-10pt}
\end{figure}

\begin{figure*}[t]
\begin{center}
   \includegraphics[width=1\linewidth]{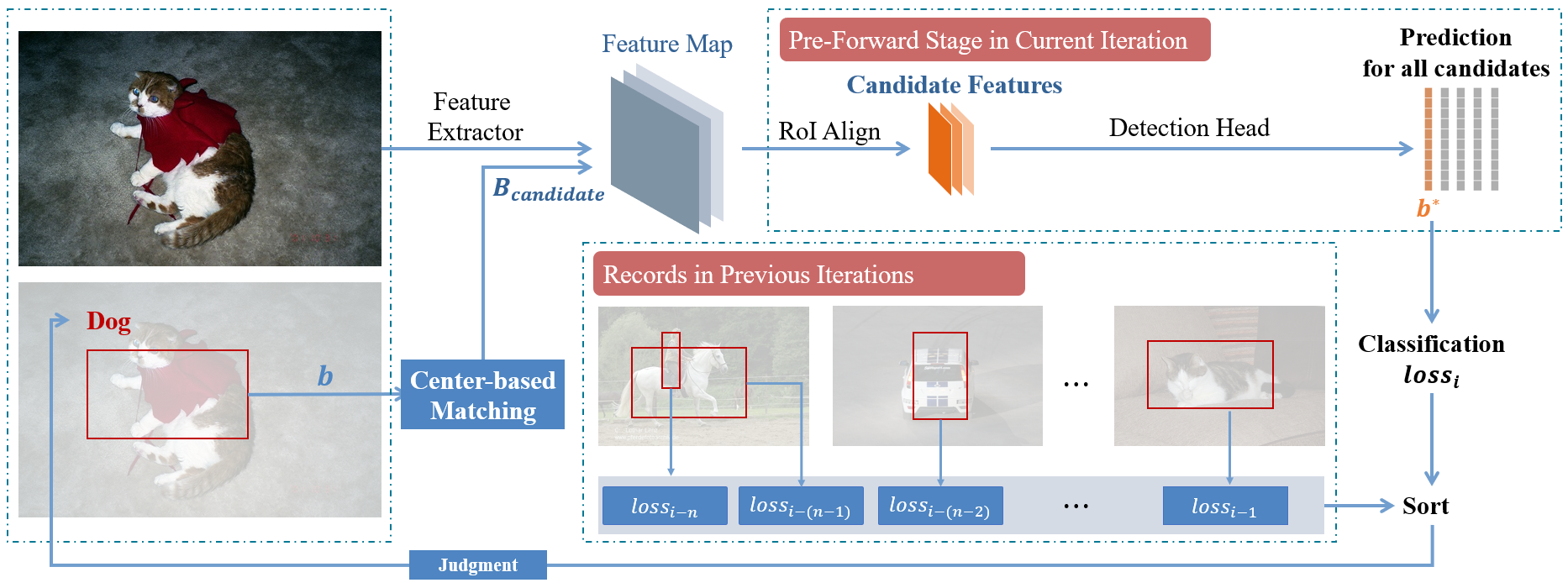}
\end{center}
\vspace{-10pt}
   \caption{Architecture of the cross-iteration noise judgment. A dynamic threshold is calculated according to the loss values maintained in the previous iterations. The CINJ judges the credibility of the current labels by comparing their loss values with the threshold. The labels judged to be clean are maintained for training, whereas the labels judged to be noisy are modified, as described in Section~\ref{sec:pre-forward}.}
\label{fig:CINJ}
\vspace{-5pt}
\end{figure*}

\vspace{-10pt}
\subsection{Center-Matching Correction}
\label{sec:CM}
Previous works~\cite{tanaka2018joint,mao2020noisy} verified the conclusion that the noise level of classification labels and location annotations is positively correlated with the corresponding training loss and proposed that modifying the annotations in the direction of reducing the loss value can reduce the noise level of the dataset. Inspired by their work, we also attempt to correct the training annotation with prediction and optimize the network parameters with SGD alternately. 

Because the classification head $\mathcal{H}_c$ and the regression head $\mathcal{H}_r$ are dependent on the region proposals generated by RPN $\mathcal{R}$, we decide to find the correct region proposals to deal with localization noise first. Fortunately, we notice that both the training and the inference of the RPN are class-agnostic, which implies that we can ignore the class label noise and correct the location annotation by matching the proper proposals to each annotated bounding box. 

The most straightforward method is to assign the proposals to the closest annotated bounding boxes based on their IoU. However, when the annotated bounding boxes are significantly distorted owing to noise, these bounding boxes cannot be matched to any proposals, even if the proposals are accurate. To solve this problem, we propose a center matching strategy that is more effective than the IoU, as it can match a bounding box with nearby proposals, even if the annotated bounding box is severely distorted in shape and has low IoUs with the corresponding proposals. 

We introduce the criterion that calculates the fitness between the bounding boxes and the proposals. Given an annotated bounding box $b$ and a proposal $p \in {P}$, where ${P}$ denotes all the proposals generated by the RPN, the fitness between $b$ and $p$ can be calculated as follows: 
\begin{equation}
\label{fitness}
    fit(b,p) = 1 - (\mathcal{D}(b,p) + \gamma \cdot \mathcal{C}(b,p)),
\end{equation}
$fit(b,p)$ measures the similarity between the annotated bounding box $b$ and the region proposal $p$. The center of the noisy annotated bounding boxes are usually still near the objects, even if their shapes are severely distorted. Therefore, we expect to match the region proposals and annotated bounding boxes on the basis of the distances of their centers. $\mathcal{D}(b,p)$ measures the relative distance between $b$ and $p$. Let $(x_c, y_c)$, $W$, and $H$ denote the coordinates of the center point, width, and height of the bounding box, respectively. The calculation of $\mathcal{D}(b,p)$ is formulated as follows:
\begin{equation}
\label{distance}
    \mathcal{D}(b,p) = \frac{\sqrt{(x_c^b - x_c^p)^2 + (y_c^b - y_c^p)^2}}{W_p + H_p}.
\end{equation}
Matching based on distance makes it more difficult to find the corresponding proposals for large objects than for small objects. Therefore, we divide the Euclidean distance by the size of the object to mitigate the difference between objects of different sizes. For simplicity, the size of the objects are approximately represented by the half circumference of the proposals.
$\mathcal{C}(b,p)$ denotes the size difference between bounding box $b$ and proposal $p$ as follows:

\begin{equation}
\label{size_cost}
    \mathcal{C}(b,p) = \left\vert \frac{W_b + H_b}{W_p + H_p} - 1 \right\rvert.
\end{equation}

$\mathcal{C}(b,p)$ is designed to prevent proposals from being assigned to incorrect overlapping bounding boxes. We expect this term to work only when the differences in the calculated $\mathcal{D}(b,p)$ are not significant. Therefore, the parameter $\gamma$ is set to $0.1$ to prevent this term from interfering with the distance determination in the usual case.

For the proposals ${P}_h $ that have Top-100 objectness scores generated by the RPN, where ${P}_h \subset {P}$, we calculate the fitness between the proposals and the annotated bounding box $b$, and maintain the proposals with a large fitness as follows:

\begin{equation}
\label{assigned proposal}
    B_{candidate} = \{~p~|~fit(b,p)>0.9, p \in {P}_h~\}.
\end{equation}

Subsequently, a few proposals in $B_{candidate}$ that have Top-10 objectness scores are maintained and the others are discarded to reduce redundancy. We use the proposal that has the highest objectness score $p^*$ to perform class-agnostic refinement on each annotated bounding box $b$ as follows:
\begin{equation}
\label{eq:step_1}
    b^* = \alpha \cdot p^* + (1-\alpha) \cdot b
\end{equation}

After all, the center-matching correction initially modifies the annotated bounding box $b$ to $b^*$, which is used for classification noise judgment and label refinement. At the same time, other candidate regions $B_{candidate}$ are utilized for further localization refinement, as described in Section~\ref{sec:pre-forward}.

\vspace{-10pt}
\subsection{Cross-Iteration Noise Judgment}
\label{sec:CINJ}

We noticed that the correctness of the classification labels is a dominant factor in the magnitude of the classification loss, as shown in Figure~\ref{fig:distribution}. This phenomenon motivate us to distinguish the incorrect labels from the correct labels, which allow us to refine the noisy labels and maintain the clean class labels. Thus,We propose a noise judgment method called cross-iteration noise judgment (CINJ), which uses an adaptive threshold to judge the classification labels of the current training iteration, as shown in Figure~\ref{fig:CINJ}. 

We maintain the classification loss $\mathcal{L}(b^*,y)$ for all the annotated bounding boxes in the pre-forward stage of each iteration. The CINJ adaptively discriminates the credibility of the current class labels by comparing the relative magnitude of the classification losses in the current iteration with the loss values maintained in the previous iterations.

Let $\mathcal{L}_i(b^*,y)$ denote the classification loss corresponding to an annotation $\{b,y\}$ in the $i$-th iteration. The CINJ mechanism maintains a queue $\mathcal{Q} = [\mathcal{L}_{i-N}, \mathcal{L}_{i-N+1},...,\mathcal{L}_{i-1} ]$ of loss values, which contains $N$ loss values in the latest dozens of iterations. For simplicity, we assume that there is only one image in each iteration. Each $\mathcal{L}_j \in \mathcal{Q}$ uniquely corresponds to an annotation $\{b_j,y_j\}$ in the training dataset. Let $\mathcal{Q}^*= [\mathcal{L}^*_{1}, \mathcal{L}^*_{2},...,\mathcal{L}^*_{N} ]$ denote the result of sorting $\mathcal{Q}$ in the ascending order. When $R\%$ of the classification labels in the dataset are incorrect and $\mathcal{Q}^*$ is sufficiently long, we can assume that approximately $R\%$ of the loss values in $\mathcal{Q}^*$ correspond to the incorrect labels. Because the noisy labels cause larger loss values than those of correct labels, as shown in Figure~\ref{fig:distribution}, the loss values corresponding to the $R\%$ incorrect labels are clustered at the back end of $\mathcal{Q}^*$. Therefore, we use an acceptance rate $r = 1 - R\%$ to compute an adaptive threshold $\mathcal{T} = \mathcal{Q}_{\lfloor r \cdot N \rfloor}^*$, and labels with loss values greater than this threshold $\mathcal{T}$ are considered as noisy labels. As queue $\mathcal{Q}$ is updated after each iteration, $\mathcal{T}$ dynamically changes.

\begin{figure}[t]
\begin{center}
   \includegraphics[width=1\linewidth]{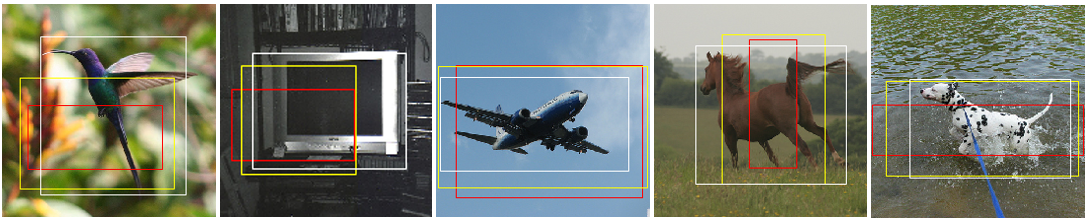}
\end{center}
\vspace{-10pt}
   \caption{Visualization of the bounding box refinement during training. The red boxes denote the noisy annotations in the training dataset with the noisy parameters of $N_{Label} = 40\%$ and $N_{BBox} = 40\%$. The correction results of the first step are shown in yellow boxes, whereas the final correction results are shown in white boxes.}
\label{fig:vis}
\end{figure}

\begin{table*}[t]
  \begin{center}
    \setlength{\tabcolsep}{1mm}{
  \begin{tabular}{l|ccc|cccc|cccc}
    \hline
     $N_{BBox}$  & \multicolumn{3}{c|}{$0\%$} & \multicolumn{4}{c|}{$20\%$} & \multicolumn{4}{c}{$40\%$}\\
    \hline
     $N_{Label}$  & $20\% $ & $40\% $ & $60\%$ & $0\%$ & $20\% $ & $40\% $ & $60\%$ & $0\%$ & $20\% $ & $40\% $ & $60\%$ \\
    \hline
    \hline
    Baseline \cite{ren2015faster,chen2019mmdetection}       &  $74.2$ &  $70.3$ & $65.8$ &  $75.5$ &  $73.1$ & $69.6$ &  $63.2$ &  $58.9$ & $55.9$ &  $53.7$ &  $47.9$ \\
    Co-teaching \cite{sugiyama2018co,li2020towards}    &  $76.5$ &  $74.1$ & $69.9$ &  $75.6$ &  $73.2$ & $69.7$ &  $65.1$ &  $60.6$ & $59.7$ &  $55.8$ &  $50.4$ \\
    NOTE-RCNN \cite{gao2019note,li2020towards}    &  $76.7$ &  $74.9$ & $69.9$ &  $76.0$ &  $73.7$ & $70.1$ &  $65.8$ &  $63.4$ & $61.5$ &  $57.8$ &  $53.7$ \\
    CA-BBC \cite{li2020towards}       &  $79.1$ &  $77.7$ & $74.1$ &  $77.9$ &  $76.7$ & $74.8$ &  $71.9$ &  $71.9$ & $70.6$ &  $69.1$ &  $64.5$ \\
    \hline
    \hline
    Baseline + BR     &  $-$ &  $-$ & $-$ &  \bm{$78.4$} &  $74.3$ & $71.2$ &  $66.0$ &  \bm{$73.4$} & $70.2$ &  $65.9$ &  $60.4$ \\
    Baseline + LR    &\bm{$79.9$}&\bm{$79.3$}&\bm{$76.6$} &  $-$ &  $76.9$ & $75.5$ &  $71.6$ &  $-$ & $60.3$ &  $58.6$ &  $55.0$ \\
    Proposal          &\bm{$79.9$}&\bm{$79.3$}&\bm{$76.6$}&\bm{$78.4$}&\bm{$77.2$}&\bm{$76.4$}&\bm{$72.8$} &\bm{$73.4$}&\bm{$71.7$}&\bm{$70.3$}&\bm{$66.6$} \\
    \hline
  \end{tabular}}
\end{center}
\caption{Comparison with state-of-the-art methods on the noisy PASCAL VOC dataset 07+12. The results (mAP@50) are evaluated on the clean test split of the VOC 2007 dataset.Bounding box refinement(BR) includes the initial bounding box correction in the center-matching correction (Eq.~\ref{eq:step_1}) and the final correction (Eq.~\ref{eq:step_2}). Label refinement(LR) includes the CINJ (Section~\ref{sec:CINJ}) and the class label refinement (Eq.~\ref{eq:pseudo_1}).}
\label{tab:pascal}
\vspace{-10pt}
\end{table*}

\begin{table*}[t]
  \begin{center}
    \setlength{\tabcolsep}{2mm}{
  \begin{tabular}{l|cc|cc}
    \hline
     \multirow{2}*{Method}  & \multicolumn{2}{c|}{$N_l = 20\%, N_b = 20\%$} & \multicolumn{2}{c}{$N_l = 40\%, N_b = 40\%$}\\
    \cline{2-5}
     ~ & mAP@0.5 & mAP@[.5,.95]& mAP@0.5& mAP@[.5,.95]\\
    \hline
    \hline
    Baseline \cite{ren2015faster,chen2019mmdetection}     & $47.9$ & $23.9$ & $29.7$ & $10.3$ \\
    Co-teaching \cite{sugiyama2018co,li2020towards} & $49.7$ & $24.6$ & $35.9$ & $14.6$ \\
    NOTE-RCNN \cite{gao2019note,li2020towards}  & $50.4$ & $25.1$ & $38.5$ & $15.2$ \\
    CA-BBC \cite{li2020towards}    & $53.5$ & $27.7$ & $47.4$ & $21.2$ \\
    \hline
    \hline
    Baseline + BR         & $52.9$     &$28.1$     & $44.6$      & $19.3$ \\
    Baseline + LR         & $54.0$     &$29.7$     & $40.1$      & $17.2$ \\
    Proposal              & \bm{$54.3$}&\bm{$30.2$}& \bm{$47.5$} & \bm{$21.7$} \\
    \hline
  \end{tabular}}
\end{center}
\caption{Comparison with state-of-the-art methods on the MS-COCO dataset. The results are evaluated on the clean validation split of the MS-COCO 2017.}
\label{tab:coco}
\vspace{-15pt}
\end{table*}

\vspace{-10pt}
\section{Experiments}
\subsection{Dataset}

We manually introduce different levels of symmetric classification label noise and uniform localization noise into the clean PASCAL VOC~\cite{everingham2010pascal} and MS-COCO~\cite{lin2014microsoft} datasets. We use two parameters, $N_{Label}$ and $N_{BBox}$, to control the level of  symmetric class label noise and uniform bounding box localization noise, respectively. In particular, for all artificial noisy datasets, $N_{Label}$ of the classification labels are randomly chosen and replaced by other categories. The horizontal coordinates of all bounding boxes are perturbed by a number of pixels uniformly drawn from $[-wN_{BBox}, +wN_{BBox}]$, and from $[-hN_{BBox}, +hN_{BBox}]$ for the vertical coordinates, where $w$ and $h$ denote the width and height of the corresponding box, respectively. By combining these two parameters, we obtain artificial noisy datasets with different levels of noise, where the classification noise and the localization noise are entangled together. 

\vspace{-10pt}
\subsection{Implementation}
We construct the proposed framework based on a Faster R-CNN with ResNet-50 \cite{he2016deep} and FPN~\cite{lin2017feature} as the feature extractor. All the models trained on the noisy PASCAL VOC datasets use an SGD optimizer with a learning rate of 0.01, a momentum of 0.9, and a weight decay of $1 \times 10^{-4}$. For the experiments on the MS-COCO dataset, the learning rate is changed to $5 \times 10^{-3}$, and the others remain unchanged. Our implementation is based on the mmdetection toolbox~\cite{chen2019mmdetection} and the training schedule basically follow the default training configuration provided by the toolbox. When training the Faster R-CNN using our proposal, we start the judgment and correction after the first epoch to obtain stable results. The length $N$ for queue $\mathcal{Q}$ is set to $128$, and the acceptance rate $r$ is set close to the percentage of correct class labels in the training dataset for all the results reported in Table~\ref{tab:pascal}. $\alpha$ is set to $0.2$ and $0.4$ for datasets with noise rates $N_{BBox} = 20\%$ and $40\%$, respectively. The quantitative sensitive analysis on the hyperparameters are shown in the Supplementary Material.

\vspace{-10pt}
\subsection{Experiment Results}
We apply our proposal on the PASCAL VOC 07+12 dataset with different levels of artificial noise, and the results are shown in Table~\ref{tab:pascal}. Our proposed method outperforms state-of-the-art methods by a significant margin on all noisy datasets with different noise levels. Figure~\ref{fig:vis} shows the results of the bounding box correction while training the detector on a dataset with $N_{Label} = 40\%, N_{BBox} = 40\%$. Our proposed method can significantly reduce localization noise even if the classification labels are unreliable. Furthermore, center-matching correction can correct the bounding boxes with high-level noise. The evaluation of the corrected training dataset is provided in Supplementary Material.

Table~\ref{tab:coco} shows the experimental results on the noisy MS-COCO dataset. Our proposal significantly outperforms state-of-the-art methods under the supervision of noisy datasets. The experimental results confirm that our proposal is effective in alleviating the negative impact caused by noisy annotations.

\begin{table}[t]
  \begin{center}
    \setlength{\tabcolsep}{5mm}{
  \begin{tabular}{cc||cc}
    \hline
     \multicolumn{2}{c||}{Bounding Box Refinement (BR)}  & \multicolumn{2}{c}{$N_{BBox}$}    \\
    \hline
     \makecell[c]{Center-Matching\\ (Eq.~\ref{eq:step_1}) }      &  \makecell[c]{Refinement \\(Eq.~\ref{eq:step_2})}         & $20\%$     & $40\%$  \\
     \hline
    ~          & ~                & $75.5$     & $58.9$  \\
    \checkmark & ~                & $77.3$     & $68.4$  \\
    \checkmark &\checkmark        & \bm{$78.4$}& \bm{$73.4$}  \\
    \hline
  \end{tabular}}
\end{center}
\caption{Ablation experiments: mAP of the noisy localization annotation refinement.}
\label{tab:CM_ablation}
\end{table}

\begin{table}[t]
  \begin{center}
    \setlength{\tabcolsep}{3mm}{
  \begin{tabular}{cc||ccc}
    \hline
     \multicolumn{2}{c||}{Label Refinement (LR)}  & \multicolumn{3}{c}{$N_{Label}$}    \\
    \cline{1-5}
     CINJ        & Refinement (Eq.~\ref{eq:pseudo_1})       & $20\%$     & $40\%$    & $60\%$   \\
     \hline
    ~          & ~                & $74.2$     & $70.3$    & $65.8$    \\
    \checkmark & ~                & $78.3$     & $74.4$    & $70.0$    \\
    ~          &\checkmark        & $78.9$     & $76.9$    & $71.4$    \\
    \checkmark &\checkmark        &\bm{$79.9$} &\bm{$79.3$}&\bm{$76.6$}   \\
    \hline
  \end{tabular}}
\end{center}
\caption{Ablation experiments: mAP of the noisy label refinement.}
\label{tab:cinj_ablation}
\end{table}

\vspace{-10pt}
\subsection{Ablation Study}
In this section, we conduct ablation experiments on noisy Pascal VOC datasets to examine the effectiveness of each part of our proposal. As each of our modules explicitly targets a particular type of noise, we separately examine our methods on the datasets with the corresponding noise. 

For the bounding box noise, the bounding box refinement (BR) involves two steps. The first step is performed by the center-matching correction and initially corrects the bounding boxes regardless of the class labels. The second step uses the class labels to choose candidates with top-2 confidence in the corresponding category and subsequently correct the corresponding bounding box using Eq.~\ref{eq:step_2}. Table~\ref{tab:CM_ablation} shows the effect of the two steps, both of which contribute to the performance of our proposal.

For the classification noise, our label refinement (LR) first performs the CINJ to judge the correctness of the labels and subsequently replace those labels judged to be noisy with pseudo-labels. Table~\ref{tab:cinj_ablation} shows the effectiveness of both steps when dealing with noisy class labels. The results in the first row are obtained using the vanilla Faster R-CNN. The second row shows the results when only the CINJ is performed, and all labels judged to be noisy are discarded without modification. The third row shows the results when all class labels are replaced by high-confidence pseudo-labels without noise judgment. When the two steps work together, the proposed method achieves the best performance.

\begin{table}[t]
  \begin{center}
  \begin{floatrow}
 \setlength{\tabcolsep}{2mm}{
 \capbtabbox{
  \begin{tabular}{l||ccc}
    \hline
    \multirow{2}*{Method}  & \multicolumn{3}{c}{Pair-Noise Rate}    \\
    \cline{2-4}
    ~  & $20\%$ & $30\%$ & $40\%$     \\
    \hline
    Baseline        & $76.3$     & $71.3$    & $61.7$   \\
    Baseline + LR   & \bm{$79.8$}     & \bm{$78.6$}    & \bm{$69.8$}   \\
    \hline
  \end{tabular}{
\caption{mAP on datasets with pair classification label noise. }
\label{tab:pair}}}

\capbtabbox{
  \begin{tabular}{l||ccc}
    \hline
    \multirow{2}*{Method}  & \multicolumn{3}{c}{$\sigma^2$}    \\
    \cline{2-4}
    ~  & $0.2$ & $0.25$ & $0.3$     \\
    \hline
    Baseline        & $71.5$     & $66.4$    & $59.8$   \\
    Baseline + BR   & \bm{$74.9$}     & \bm{$73.1$}    & \bm{$71.0$}   \\
    \hline
  \end{tabular}}{
\caption{mAP on datasets with Gaussian localization annotation noise.}
\label{tab:Gaussian}}}

\end{floatrow}
\end{center}
\vspace{-15pt}
\end{table}

\vspace{-10pt}
\subsection{Evaluation on Other Types of Noise}
We further construct more challenging noisy datasets to examine the effectiveness of our proposal. For the classification labels, we introduce pair noise, which randomly select $R\%$ samples from each class $y$ and change their label indices to $y+1$. For the localization annotations, we introduce Gaussian noise. The horizontal coordinates of all bounding boxes are perturbed by $\delta{w}$ pixels and by $\delta{h}$ for the vertical coordinates, where $w$ and $h$ denote the width and height of the corresponding box, respectively, and $\delta$ is drawn from a Gaussian distribution $\delta \sim N(0,\sigma^2)$. The precise definitions of the noises are provided in the Supplementary Material.

The results are shown in Tables~\ref{tab:pair} and~\ref{tab:Gaussian}. The experimental results confirm that our proposal can significantly alleviate the negative impact caused by noisy annotations, despite the type of noise.

\vspace{-10pt}
\section{Conclusion}
\vspace{-5pt}
In this study, we propose a novel problem setting of training object detector on datasets with entangled classification noise and localization annotation noise. We proposed an effective correction method based on center matching for localization annotation noise to match extremely noisy bounding boxes to the proposals generated by the RPN. We observed the behavior of the object detector under the supervision of noisy class labels and noticed that noisy labels led to significantly high values of classification loss. On the basis of this finding, we proposed a noise-judging method, called cross-iteration noise judgment, to identify incorrect training labels. We arrange the steps with an additional pre-forward stage to decouple the entangled noise. Our proposal significantly outperforms the baseline by reducing the noise level of the training dataset. Each proposed step is experimentally shown to be effective in alleviating the negative impact caused by incorrect annotations.

\vspace{-10pt}
\subsubsection*{ACKNOWLEDGEMENTS} 
This work is partially supported by JST CREST 90249938 and JSPS KAKENHI 18H03254.

\bibliography{egbib}
\end{document}


\maketitle

\section{Definition of noise}
\subsection{Symmetry Label Noise}
The definition of transition matrix $\mathcal{Q}$ of symmetric noise is as follow, 

\begin{equation}
    \mathcal{Q} = 
    {
    \left[
    \begin{matrix}
    1-r           & \frac{r}{n-1} &\cdots         & \frac{r}{n-1}  & \frac{r}{n-1} \\
    \frac{r}{n-1} & 1-r           & \frac{r}{n-1} & \cdots         & \frac{r}{n-1} \\
    \vdots        & ~             & \ddots        & ~              &\vdots\\
    \frac{r}{n-1} &\cdots         &\frac{r}{n-1}  & 1-r            &\frac{r}{n-1} \\
    \frac{r}{n-1} &\frac{r}{n-1}  &\cdots         &\frac{r}{n-1}   & 1-r
    
    \end{matrix}   
    \right]
    }
\end{equation}
where $r$ is the noise rate and $n$ is number of the class.

\subsection{Pair Label Noise}
The definition of transition matrix $\mathcal{Q}$ of pair noise is as follow, 
\begin{equation}
    \mathcal{Q} = 
    {
    \left[
    \begin{matrix}
    1-r           & r             &0         & \cdots   & 0 \\
    0             & 1-r           & r        & \cdots   & 0 \\
    \vdots        & ~             & \ddots        & \ddots              &\vdots\\
    0             &\cdots         &~            & 1-r            &r \\
    r             &0  &\cdots         &0   & 1-r
    
    \end{matrix}   
    \right]
    }
\end{equation}
where $r$ is the noise rate.

\subsection{Uniform Localization Noise}
 If the original bounding boxes are presented as $[x_1,y_1,x_2,y_2]$, the noisy bounding boxes $[x_1',y_1',x_2',y_2']$ can be formulated as follows:

\begin{equation}
\label{eq:uniform}
\begin{cases}
    \vspace{5pt}
    x_1' =  x_1 + \delta_1( x_2 - x_1 ),\\
    \vspace{5pt}
    x_2' =  x_2 + \delta_2( x_2 - x_1 ),\\
    \vspace{5pt}
    y_1' =  y_1 + \delta_3( y_2 - y_1 ),\\
    \vspace{5pt}
    y_2' =  y_2 + \delta_4( y_2 - y_1 ),
\end{cases}
\end{equation}
where $\delta_i \sim U(-N_{BBox},N_{BBox})$.

\subsection{Gaussian Localization Noise}
We use the same way as above to create the Gaussian noise dataset. The difference is that the $\delta_i \sim N(0,\sigma^2)$ of Gaussian noise datasets follow the Gaussian distribution.

\section{Implementation Details}
Although our explanation about the proposal assumes that there is only one object in each image and the training batch size is 1, our experiments train the Faster R-CNN with batch size as 2. In practice, we apply the process we described in Section~3 to each image and each object. During the pre-forward stage, each annotated object will generate one corresponding classification loss. We record the value of the loss and then judge their correctness and refine them separately, as shown in Algorithm.~\ref{alg:1}.

\begin{algorithm}[t]
\label{alg:1}
  \caption{Overall Architecture} 
 
    \textbf{Input:} Image $\mathcal{X}$, noisy annotation $\{\mathcal{Y},\mathcal{B}\}$.\;
    $\mathcal{Q} \leftarrow \{\infty\}^{N}$\;
    \While{not MaxIters}{
    \For {$x \in$ Batch}
    {Image $x$ with annotation $Y = \{y_i\}, B = \{b_i\}$\;
     Generate proposals ${P}$ by RPN$(x)$\;
    \For{$\{y,b\}\in \{Y,B\}$}{
     $b^*, {P}_{b} \leftarrow$  Center-Matching$(b, {P})$\;
    \For{$b_i \in \{b^*\} \cup {P}_{b}$}{
      Calculate score $p(x,b_i)$ by Pre-Forward\;
      $b_i^r \leftarrow$ Regress$(b_i)$ by Pre-Forward\;
     ${B }_{r} \leftarrow {B }_{r} \cup \{b_i^r\} $\;
    }
     Calculate $b^*_m$ by Eq.~(14)\;
  
     Calculate classification loss $\mathcal{L}(b^*, y)$\;
  
     $clean(y) \leftarrow$ CINJ$(\mathcal{L}(b^*), \mathcal{Q})$\;
  \eIf {$clean(y)$}{
     $Y^* \leftarrow Y^* \cup \{y\}, B^* \leftarrow B^* \cup \{b^*_m\}$\;
  }{
     Pseudo-label $y^* \leftarrow \argmax_{c}{p(c|x,b^*)}$\;
  \If{$p(y^*|x,b^*) > \sum_{c\neq{y^*}}{p(c|x,b^*)}$}{
     $Y^* \leftarrow Y^* \cup \{y^*\}, B^* \leftarrow B^* \cup \{b^*_m\}$\;
  }}
 
     $\mathcal{Q} \leftarrow \mathcal{Q} + \{\mathcal{L}(b^*)\}$\;
     $\mathcal{Q} \leftarrow \mathcal{Q} - \mathcal{Q}_0$\;
  }
     Train model with updated annotation $\{Y^*,B^*\}$\;
  }}
\end{algorithm}

\begin{table}
  \begin{center}
  \setlength{\tabcolsep}{4mm}{
  \begin{tabular}{c|cccc|c}
    \hline
    \multirow{2}*{$N_{Label}$}  & \multicolumn{4}{c|}{$N$} & \multirow{2}*{Baseline} \\
    \cline{2-5}
    ~ & 64 & 128 & 258 & 512 & ~ \\
    \hline
    \hline
    $20\%$ & $79.6$ &  \bm{$79.9$} &  $79.7$ &  $79.8$ &  $75.2$\\
    $40\%$ & $78.8$ &  $79.3$ &  \bm{$79.6$} &  $79.3$ &  $71.3$\\
    $60\%$ & $76.3$ &  \bm{$76.6$} &  $76.3$ &  $76.4$ &  $66.8$\\
    \hline
  \end{tabular}}
\end{center}
\caption{mAP of Faster R-CNN trained on datasets with classification label noise by different $N$.}
\label{tab:n}
\end{table}

\begin{table}
  \begin{center}
  \setlength{\tabcolsep}{4mm}{
  \begin{tabular}{c|ccccc|c}
    \hline
    \multirow{2}*{$N_{BBox}$}  & \multicolumn{5}{c|}{$\alpha$} & \multirow{2}*{Baseline} \\
    \cline{2-6}
    ~ & 0.1 & 0.2 & 0.3 & 0.4 & 0.5 & ~ \\
    \hline
    \hline
    $20\%$ & $78.0$ &  \bm{$78.4$} &  $77.9$ &  $77.4$ &  $77.8$ &  $75.6$\\
    $40\%$ & $72.3$ &  $73.2$ &  \bm{$73.4$} &  $73.1$ &  $72.8$ &  $58.9$\\
    \hline
  \end{tabular}}
\end{center}
\caption{mAP of Faster R-CNN trained on datasets with localization annotation noise by different $\alpha$.}
\label{tab:alpha}
\end{table}

\section{Analysis on Hyper-parameters}

There are three hyperparameters need to be determined for our proposal. They are the weight $\alpha$ for the initial bounding box correction in Eq.~(19), length $N$ for queue $\mathcal{Q}$, and the acceptance rate $r$ for CINJ. 

The acceptance rate $r$ should be close to the percentage of correct class labels in the training dataset, which can be estimated by using a small amount of training sample. However, when the dataset is extremely noisy and the dataset is unbalance, setting an acceptance rate close to the percentage of correct class labels may prevent the detector from learning hard classes. In such case, slightly increase the acceptance rate can allow the detectors to learn from hard classes. In our experiments, the acceptance rate $r$ for datasets with $N_{label}$ as $20\%,40\%$ and $60\%$ are set as $80\%,60\%$ and $50\%$.

The quantitatively sensitive analysis on $N$ is shown on Table.~\ref{tab:n}. The performance of our proposal is not sensitive to $N$, all of experiments with $N$ larger than 128 achieve relatively close results. $\alpha$ is the only tunable parameter of our proposal. The quantitatively sensitive Analysis on $\alpha$ is shown on Table.~\ref{tab:alpha}. Tuning $\alpha$ from 0.2 to 0.4 can achieve an optimal performance, and all setting of $\alpha$ from 0.1 to 0.5 significantly outperform the baseline.

\section{Analysis on Thresholds}

There are 2 thresholds need to be determined for our proposal. They are the threshold  $\mathcal{T}_{CM}$ of matching region proposals to annotated bounding boxes in Eq.(18), and the label refinement threshold $\mathcal{T}_{refine}$ in Eq.(13). The quantitatively sensitive analysis on $\mathcal{T}_{CM}$ and $\mathcal{T}_{refine}$ are shown on Table.~\ref{tab:tcm} and Table.~\ref{tab:tre}, respectively. The performance of our proposal is not sensitive to both of these two thresholds, all of the experiments achieve relatively close results.

\begin{table}  
\begin{center}  
\setlength{\tabcolsep}{4mm}{  
\begin{tabular}{c|cccc|c}   
\hline   
\multirow{2}*{$N_{Label}$}  & \multicolumn{4}{c|}{$\mathcal{T}_{CM}$} & \multirow{2}*{Baseline} \\    
\cline{2-5}    ~ & $0.8$ & $0.85$ & $0.9$ & $0.95$ & ~ \\    
\hline    
\hline    
$20\%$ & $79.6$ &  \bm{$79.9$} &  $79.7$ &  $79.8$ &  $75.2$\\    
$40\%$ & $78.8$ &  $79.3$ &  \bm{$79.6$} &  $79.3$ &  $71.3$\\    
$60\%$ & $76.3$ &  \bm{$76.6$} &  $76.3$ &  $76.4$ &  $66.8$\\    
\hline 
\end{tabular}}
\end{center}
\caption{mAP of Faster R-CNN trained on datasets with classification label noise by different $\mathcal{T}_{CM}$.}
\label{tab:tcm}
\end{table}

\begin{table}  
\begin{center}  
\setlength{\tabcolsep}{4mm}{  
\begin{tabular}{c|cccc|c}    
\hline    
\multirow{2}*{$N_{BBox}$}  & \multicolumn{4}{c|}{$\mathcal{T}_{refine}$} & \multirow{2}*{Baseline} \\    
\cline{2-5}    ~ & 0.4 & 0.5 & 0.6 & 0.7 &  ~ \\    
\hline    \hline    $20\%$ & $78.0$ &  \bm{$78.4$} &  $77.9$ &   $77.8$ &  $75.6$\\
$40\%$ & $72.3$ &  $73.2$ &  \bm{$73.4$} &    $72.8$ &  $58.9$\\    
\hline 
\end{tabular}}
\end{center}
\caption{mAP of Faster R-CNN trained on datasets with localization annotation noise by different $\mathcal{T}_{refine}$.}
\label{tab:tre}
\end{table}

\vspace{-10pt}
\section{Evaluation on Refined Datasets}
Our proposal alternately update the noisy annotation and the parameters of the detector. To prove the effectiveness of our noisy annotation refinement, we record the refined annotations in last training epoch and evaluate them by clean annotations. Table.~\ref{tab:loc} shows the results of the localization annotation refinement. We compare the refined localization annotation with clean dataset. When a refined bounding box has an IOU with any bounding box in clean annotation larger than 0.7, it is considered as correct localization. The first row shows Corloc of the noisy annotation, second row and last row show the CorLoc after center-matching and final correction, respectively. Table.~\ref{tab:cls} shows the results of the classification label judgment and refinement. Although severe localization annotation noise and the classification label noise are entangled together, our proposal can still significantly reduce the class label noise rate.
\begin{table}
  \begin{center}
  \setlength{\tabcolsep}{2mm}{
  \begin{tabular}{l|cccc|cccc}
    \hline
    $N_{BBox}$ & \multicolumn{4}{c|}{$20\%$} & \multicolumn{4}{c}{$40\%$} \\
    \hline
    $N_{Label}$ & $0\%$ & $20\%$ & $40\%$ & $60\%$ & $0\%$ & $20\%$ & $40\%$ & $60\%$  \\
    \hline
    \hline
    CorLoc$_{noisy}$ & $54.67$ &  $44.93$ &  $45.10$ &  $45.52$ &  $9.02$  &  $8.66$ &  $8.73$ &   $8.79$\\
    CorLoc$_{cm}$    & $57.35$ &  $57.23$ &  $56.97$ &  $57.62$ &  $15.44$ &  $15.39$&  $15.15$ &  $15.38$\\
    CorLoc$_{final}$ & \bm{$77.21$} &  \bm{$75.54$} &  \bm{$74.68$} &  \bm{$73.86$} &  \bm{$38.08$} &  \bm{$37.68$} &  \bm{$34.13$} &  \bm{$34.17$}\\
    \hline
  \end{tabular}}
\end{center}
\caption{Correct localization rate(\%) of refined noisy annotations. }
\label{tab:loc}
\vspace{-10pt}
\end{table}

\begin{table}[t]
  \begin{center}
  \setlength{\tabcolsep}{2mm}{
  \begin{tabular}{l|ccc|ccc|ccc}
    \hline
    $N_{BBox}$ & \multicolumn{3}{c|}{$0\%$} &\multicolumn{3}{c|}{$20\%$} & \multicolumn{3}{c}{$40\%$} \\
    \hline
    $N_{Label}$ & $20\%$ & $40\%$ & $60\%$ & $20\%$ & $40\%$ & $60\%$ & $20\%$ & $40\%$ & $60\%$  \\
    \hline
    \hline
    TP & $93.56$ &  $95.68$ &  $83.27$ &  $92.72$ &  $95.63$  &  $83.05$ &  $91.03$ &  $94.55$ &  $82.11$\\
    TN & $97.07$ &  $94.44$ &  $98.6$  &  $96.79$ &  $94.77$  &  $98.23$ &  $96.65$ &  $93.94$ &  $97.06$\\
    FP & $2.93$  &  $5.56$  &  $1.40$   &  $3.21$  &  $5.23$   &  $1.77$  &  $3.35$  &  $6.06$  &  $2.94$\\
    FN & $6.44$  &  $4.32$  &  $16.73$ &  $7.28$  &  $4.37$   &  $16.95$ &  $8.97$  &  $5.45$  &  $17.89$\\
    \hline
    \hline
    $N_{Label}^*$ &  \bm{ $3.48$ } &  \bm{  $6.31$ } &  \bm{  $13.97$ } &  \bm{  $4.23$ } &  \bm{  $6.49$ } &  \bm{  $13.81$} &  \bm{  $5.36$ } &  \bm{  $8.45$} &  \bm{  $16.21$}\\
    \hline
  \end{tabular}}
\end{center}
\caption{Accuracy(\%) of the class label noise refinement. $TP$ denotes the percentage of noisy labels detected to be noisy, and $TN$ denotes the percentage of clean labels judged to be clean. $FP$ and $FN$ denote the percentage of clean labels judged to be noisy and noisy labels judged to be clean, respectively. $N_{Label}^*$ denotes the noise rate of classification labels after the judgment and refinement.}
\label{tab:cls}
\end{table}

\vspace{-10pt}
\section{Loss Distribution of Pair Label Noise}

\begin{figure}
\begin{minipage}[b]{0.25\linewidth}
  \centering
  \centerline{\includegraphics[width=\linewidth]{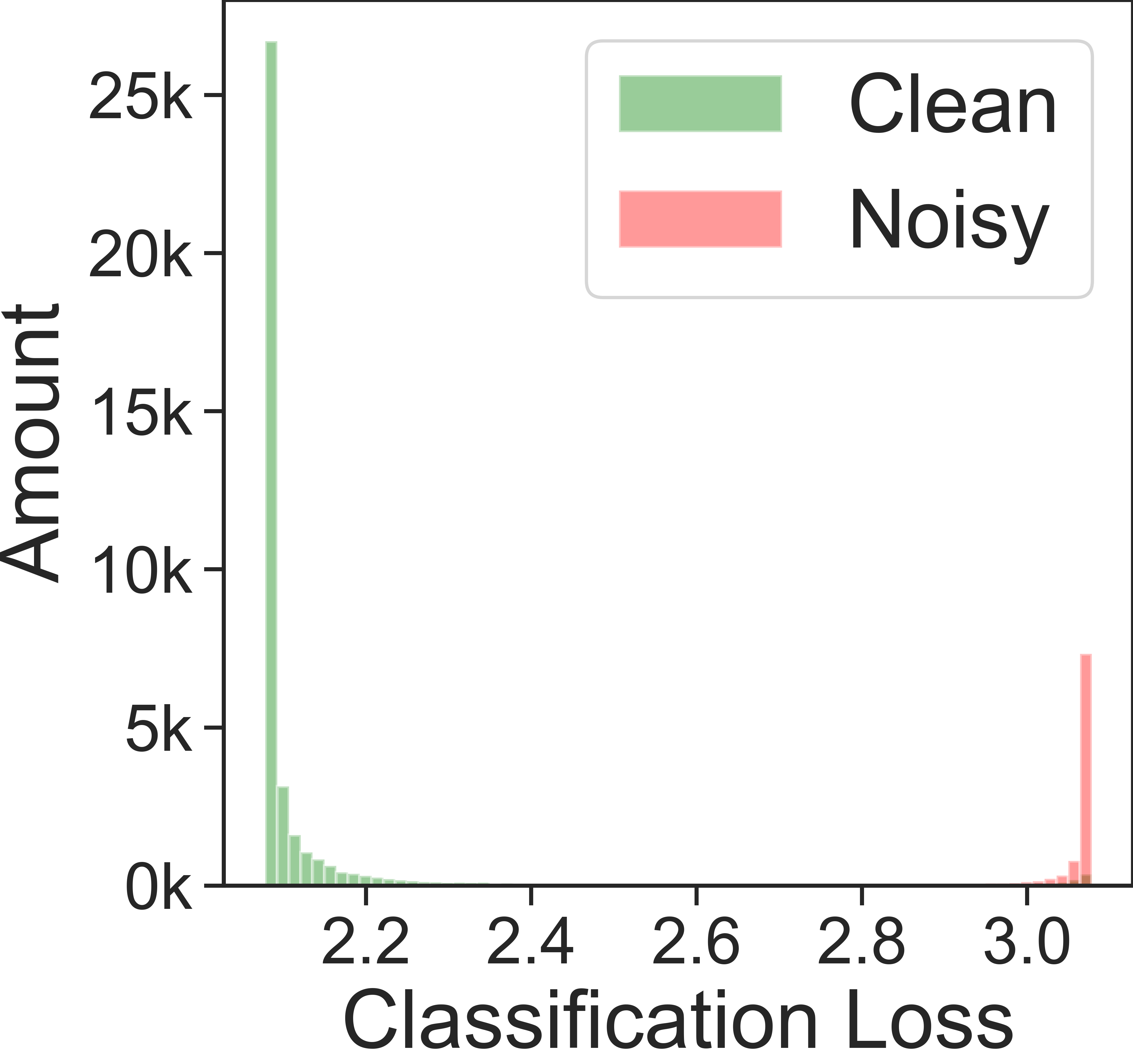}}
  \centerline{$N_{Pair} = 20\%$}
\end{minipage}
\hfill
\begin{minipage}[b]{0.25\linewidth}
  \centering
  \centerline{\includegraphics[width=\linewidth]{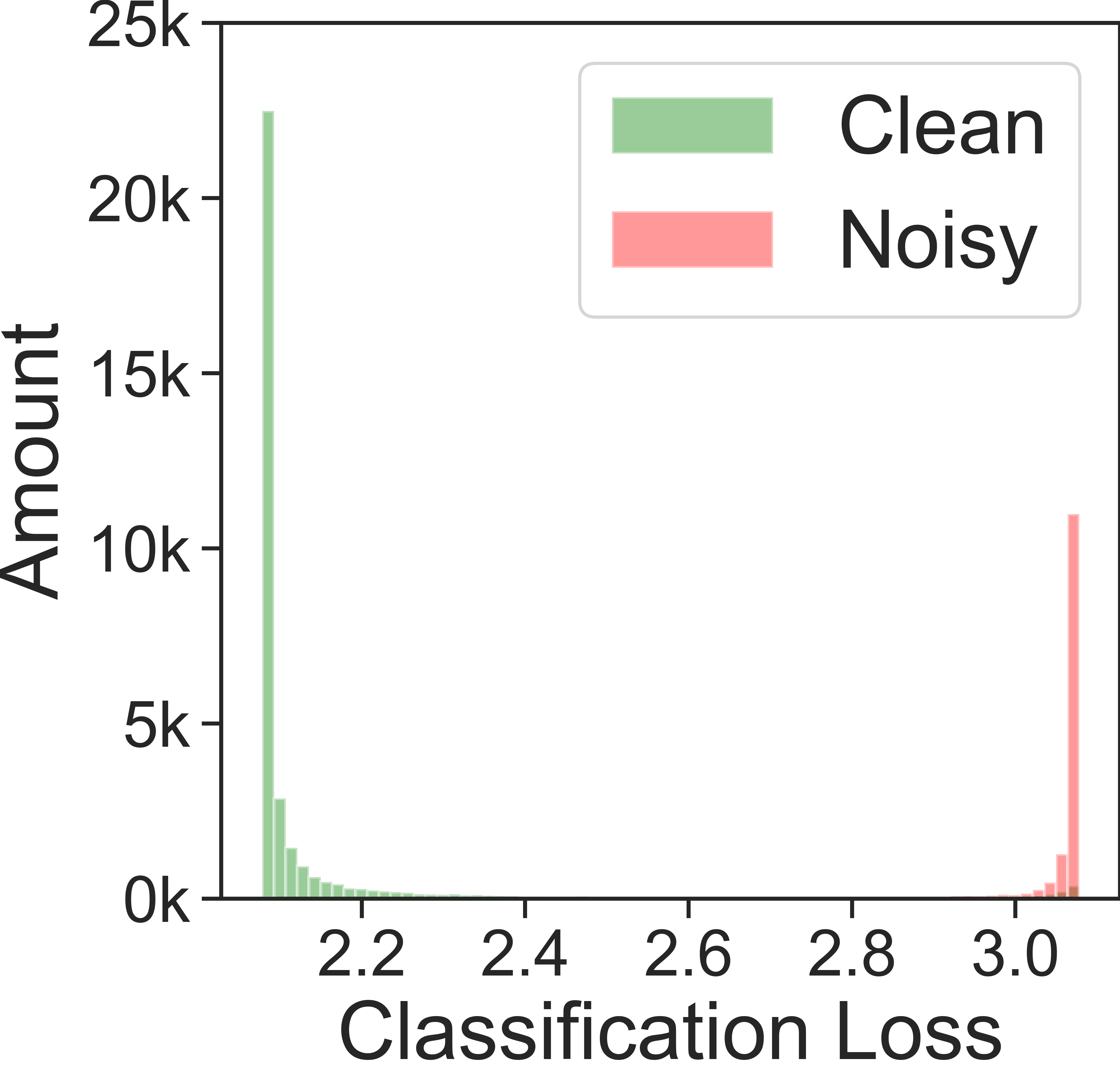}}
  \centerline{$N_{Pair} = 30\%$}
\end{minipage}
\hfill
\begin{minipage}[b]{0.25\linewidth}
  \centering
  \centerline{\includegraphics[width=\linewidth]{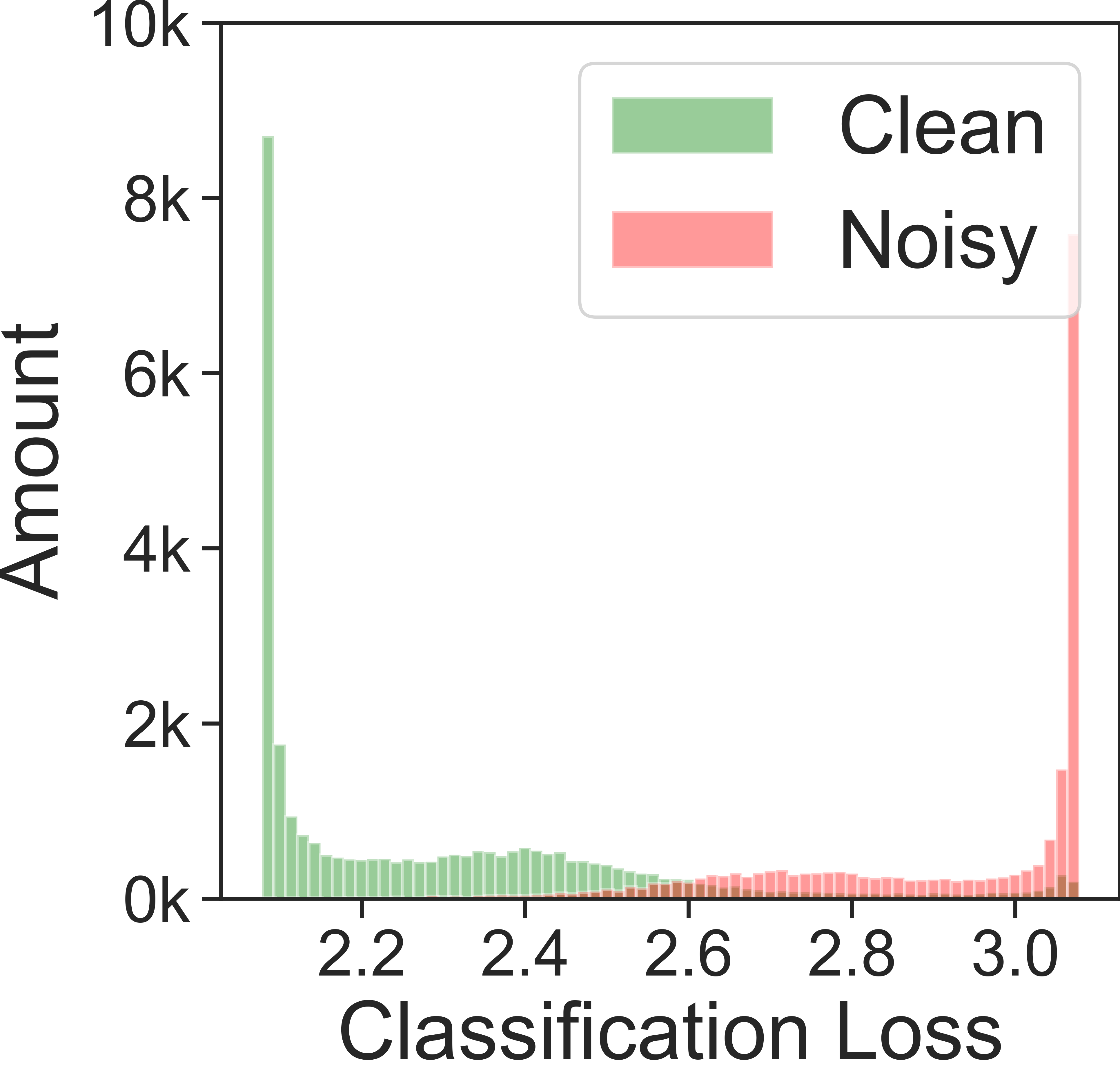}}
  \centerline{$N_{Pair} = 40\%$}
\end{minipage}

\caption{Distribution of the classification loss of datasets with pair noise.}
\label{fig:distribution}
\vspace{-10pt}
\end{figure}

Although the pair noise is much more challenging than symmetry label noise, the training losses for noisy and clean annotations are still clearly separated, as shown in Fig.~\ref{fig:distribution}. This fact ensure the effectiveness of our method on more challenging noise models such as pair noise.

